%% file: main.tex
\documentclass[10pt,twocolumn,letterpaper]{article}

\usepackage{iccv}
\usepackage{times}
\usepackage{epsfig}
\usepackage[utf8]{inputenc} 
\usepackage[T1]{fontenc}    
\usepackage{url}            
\usepackage{booktabs}       
\usepackage{amsfonts}       
\usepackage{nicefrac}       
\usepackage{microtype}      
\usepackage{multirow}
\usepackage{array}
\usepackage{tabularx}
\usepackage[table,xcdraw]{xcolor}
\usepackage[normalem]{ulem}
\useunder{\uline}{\ul}{}

\usepackage{amsmath}
\usepackage{graphicx}
\usepackage{threeparttable}
\usepackage{paralist}
\usepackage{newunicodechar}
\newunicodechar{：}{:}

\usepackage[pagebackref=true,breaklinks=true,letterpaper=true,colorlinks,bookmarks=false]{hyperref}

 \iccvfinalcopy 

\def\iccvPaperID{18} 

\ificcvfinal\fi

\begin{document}

\title{FireFly: A Synthetic Dataset for Ember Detection in Wildfire}

\author{%
Yue Hu$^{1}$ \quad Xinan Ye$^1$ \quad Yifei Liu$^{2}$ \quad Souvik Kundu$^3$ \quad Gourav Datta$^1$ \quad Srikar Mutnuri$^1$\\
Namo Asavisanu$^1$ \quad Nora Ayanian$^4$ \quad Konstantinos Psounis$^1$ \quad Peter Beerel$^1$ \quad \\
$^1$University of Southern California \quad $^2$University of California, Irvine \\
\quad $^3$Intel Labs, San Diego, USA \quad $^4$Brown University\\
\texttt{\{yhu57782,xinanye, gdatta,mutnuri,namo,kpsounis,pabeerel\}@usc.edu}\\
\texttt{yifeil23@uci.edu} \quad
\texttt{souvikk.kundu@intel.com} \quad
\texttt{nora\_ayanian@brown.edu}
}


\maketitle
\ificcvfinal\thispagestyle{empty}\fi

\begin{abstract}
This paper presents \textit{"FireFly"}, a synthetic dataset for ember detection created using Unreal Engine 4 (UE4), designed to overcome the current lack of ember-specific training resources. To create the dataset, we present a tool that allows the automated generation of the synthetic labeled dataset with adjustable parameters, enabling data diversity from various environmental conditions, making the dataset both diverse and customizable based on user requirements. We generated a total of 19,273 frames that have been used to evaluate FireFly on four popular object detection models. Further to minimize human intervention,  we leveraged a trained model to create a semi-automatic labeling process for real-life ember frames. Moreover, we demonstrated an up to 8.57\% improvement in mean Average Precision (mAP) in real-world wildfire scenarios compared to models trained exclusively on a small real dataset. 
\end{abstract}
\input{intro}

\input{dataset}
\input{experimentalEva}

\input{conclusion}

{\small
\bibliographystyle{ieee_fullname}
\bibliography{ref}
}

\end{document}

%% file: intro.tex
\section{Introduction}
Wildfires are  critical global issues, exemplified by recent events of unprecedented severity, such as the 2023 wildfires that consumed about 1.3 million acres of land and caused immense disruption~\cite{wiki:canada2023fire}. California, a region notorious for recurrent wildfires, has also experienced significant wildfire activity, with 408 fires burning a total of 131 acres by March 20, 2023~\cite{wiki:2023_California_wildfires}.

The destructive potential of wildfires extends beyond their immediate heat radiation. Airborne debris from fire fronts, known as embers or firebrands, can cause substantial damage, often outstripping the wildfires themselves~\cite{Calfire}. Research on wildland urban interface (WUI) wildfires has shown the significant role of embers in causing damage~\cite{blanchi2005investigation}.
Embers can travel substantial distances from the main fire front, making it challenging to predict their trajectories~\cite{1885-49422,IBHS2019}. They can potentially ignite fires when landing on building roofs or entering attics through vents~\cite{smith2009ember,apostol2018protect}.

\begin{figure}[!t]
    \centering
    \includegraphics[width=0.49\textwidth]{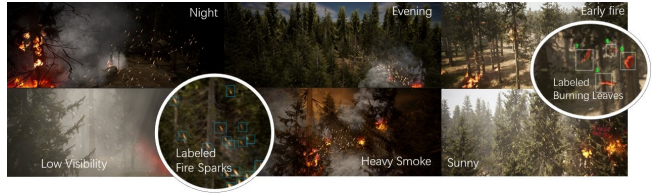}
    \caption{Example of various weather and severity conditions in our synthetic dataset (a) night, (b) evening, (c) early small fire, (d) low visibility, (e) heavy smoke and (f) sunny. The circles show the zoomed-in image with auto-generated labels.}
    \label{fig:scenes}
\end{figure}

Previous research has applied machine learning systems for wildfire detection and perimeter mapping\cite{WIFIRE,IQ,jain2020review}, which are potentially deployable across a range of platforms, such as watch towers, helikites aerostats~\cite{HELIKITE}, and autonomous drones~\cite{bbc:2023_firefighting_drone}. An ember detection system could play a crucial role after mandatory evacuations, focusing on vulnerable structures and potentially preventing severe damage. Detecting and tracking embers could guide intelligent response systems to prioritize the actions of first responders or guide drones to targeted areas for fire retardant delivery~\cite{bbc:2023_firefighting_drone,drones5010017,nationalgeographic,DSLRpros}.
However, a significant obstacle in developing such systems is the lack of datasets specifically designed for ember detection. Current datasets primarily focus on flames and smoke~\cite{data:flame,data:FireNET,data:MIVIA,data:Domestic}, leaving a gap in the ember-specific training material. Collecting real-world data from wildfires poses significant risks and does not capture the diversity of environments where embers may occur. Although manual annotation of real experiment videos has been attempted~\cite{pan2022deep}, this process is highly time-consuming. 

To address this gap, this paper presents \textit{FireFly}, a synthetic wildfire ember dataset created using Unreal Engine 4 (UE4)~\cite{UE}. Developing synthetic datasets using a game engine has multiple advantages: it significantly reduces the reliance of computer vision machine learning on manual annotations and users can readily diversify the dataset and enhance model robustness through multi-scene generation, as depicted in Fig~\ref{fig:scenes}. This is accomplished by adjusting parameters including forest type, ember type, size, season, atmospheric environment, background, lighting condition, and camera angle, thereby enriching the dataset's variability.

We evaluated the dataset using four popular object detection models~\cite{lin2017focal,peize2020sparse,wang2022yolov7,detr}, providing insights into the capabilities and limitations of these models in identifying multiple small objects in a single frame. We also performed inference using real world forest fire recordings with the models to demonstrate their ability to generalize to real-world images after training with FireFly.


This paper aims to contribute towards the development of more effective wildfire response systems by providing the synthetic dataset specifically for wildfire embers. Our contributions are as follows.
\begin{compactitem}
\item We have developed custom tools to automatically generate synthetic labeled dataset with dense small objects, and make the parameters controlling the diversity of the data adjustable.
\item Using this tool we generated the first large-scale synthetic dataset for ember detection.
Our dataset consist of 19,273 images with burning leaves and sparks spanning diverse environmental conditions such as weather, light and atmospheric visibility, forest types, and smoke 
intensity.\footnote{The dataset and associated program are open-sourced at https://github.com/ERGOWHO/Firefly2.0.git}
\item We have demonstrated that our synthetic dataset can serve as an effective tool for comparing the performance metrics of various mainstream detection models under extreme testing conditions as well as enabling a semi-automatic labelling flow of real-life ember frames. We performed empirical evaluations to demonstrate the efficiency of the presented on real world use case. Specifically, our method offers substantial enhancements in the context of real-world wildfire scenarios. We observed significant improvements in the mean Average Precision (mAP) detection metrics, with an increase of up to 8.57\% compared to models that were trained solely on a smaller real dataset.
\end{compactitem}


%% file: dataset.tex
\section{Creation of the FireFly Dataset}

\begin{table*}[!ht]
    \centering
    \footnotesize
    \caption{Firefly Dataset}
    \begin{threeparttable}
        \begin{tabular}{ccccc}
        \toprule
            Ember Type & Scene & \# Average Objects Per Frame & Ember Size & \#frames \\ 
            \midrule
            Burning Leaves & River/Dead Tree Forest$^1$ & 1 & Large & 0.9k \\ 
            ~ & Pond/Dead Tree Forest & 1 & Small/Medium & 0.7k \\ 
            ~ & Fog/Dry Wood/Dead Tree Forest & 1 & Large & 0.7k \\ \
            ~ & Fog/Dry Wood/Dead Tree Forest & 2 & Medium & 0.7k \\ 
            ~ & Fog/Dry Wood/Dead Tree Forest$^2$ & 15 & Small & 1.2k \\ 
            ~ & Sunny$^3$ & 50 & Medium/Large & 1.8k \\
            \midrule
            Sparks & Cloudy Daytime/Tree Top & 200 & Medium & 1.3k \\ 
            ~ & Evening Sunset/Heavy Smoke & 200 & Medium & 1.7k \\ 
            ~ & Mist/Low Visibility & 200 & Medium & 1.5k \\ 
            ~ & Night/Warm Temperature & 200 & Small & 0.8k \\ 
            ~ & Daytime & 200 & Small & 0.8k \\ 
            ~ & Night/Cold Temperature & 200 & Small/Medium & 2.4k \\ 
            ~ & Mix Skylight/Temperature$^4$ & 200 & Small & 2.4k \\ 
            \midrule
            Empty & Evening Sunset Time/Sunny & 0 & NA & 2.3k \\ 
        \bottomrule
        \end{tabular}
        \label{Table:Dataset}
    \end{threeparttable}
\end{table*}
\begin{figure}[!t]
    \centering
    \includegraphics[width=0.49\textwidth]{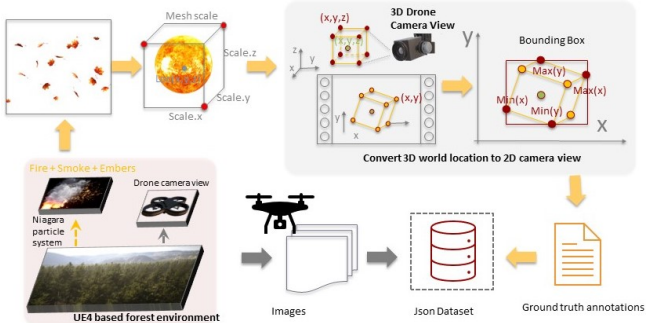}
    \caption{Workflow for generating FireFly}
    \label{fig:workflow}
\end{figure} 
The FireFly dataset is created in UE4 using two distinct forest types, dense taiga and deadwood forest, as the base environment~\cite{UEConifer}. We then adjust various parameters such as weather, light, and fog to create diverse world environments. These parameters can be dynamically adjusted or preset with a user-defined file.

The dataset comprises 14 scenes, each with a randomly selected fire point. A camera simulates the perspective of a firefighting drone, which can be manually positioned or customized via an external file. The camera uses a perspective projection mode with a 90-degree field of view and a 2560 x 1080 resolution and generates a configurable number of output frames. 

Embers in FireFly are created using UE4's Niagara particle system. We define ember data, including three-dimensional coordinates, size, and particle count, using blueprints and C++ scripts. This data is used to calculate the bounding box of each particle in the 3D world, which is then converted to a 2D bounding box based on the camera's view. To reduce distortion, we manually calibrate the ember size. 
The bounding box and id are combined with the image information captured by the camera to create the bounding box ground truth for the frames. The workflow is illustrated in Fig.~\ref{fig:workflow}.

\section{Overview of the FireFly Dataset}


As shown in Tab.~\ref{Table:Dataset}, the FireFly dataset contains 19,273 frames of data, including 16,904 positive samples with embers and 2,369 frames of negative samples without embers. According to the type of ember, the dataset is partitioned into two categories: `burning leaves' that are characterized by larger individual sizes but fewer in quantity, and `sparks' that are smaller in size but occur in denser distributions. These are two common ignition sources for forest fires. 

Burning leaves contains 6k frames of data, and its scenes include a river and pond with slight water surface reflection, and atmospheric haze environment with low visibility in autumn. The forest background was chosen as dry dead trees. The size of the burning leaves varies between large ($60\times60$ pixel average), medium ($40\times40$ pixel average), and small ($20\times20$ pixel average). When firebrands are formed and move under the action of wind, their density will drop sharply with the increase of flight distance~\cite{wang2011analysis}. Burning leaves scenes are taken from a significant distance away from the front edge of the fire, and their number is set at 1 to 50 targets per frame.

Sparks contains 10.8k frames of data, and its main scenes include day, evening, night, and low-visibility environments, with some close-ups to the fire. Most of sparks are close to the fire source, with small size, large number, and fast flying speed. There are on average 200 small detection targets per frame. We divide the dataset according to the size of the sparks into small ($3\times17$ pixel average), medium ($3\times45$ pixel average), and large ($11\times49$ pixel average).

To better capture the complexity of wildfire embers, our dataset includes a larger number of positive samples (16.7k frames) than negative ones (2.3k frames). Despite deviating from real-world prevalence of negative cases, this enables more effective training and improves model performance, particularly in reducing false negatives (underreported wildfires)

We select several representative datasets with different numbers of embers for testing. In the subsequent experiments, we will use the dataset from the River/Dead Tree Forest$^1$ scene in Table 1 as the single ember dataset in Fig.~\ref{fig:modelresults}. The dataset from the Fog/Dry Wood/Dead Tree Forest$^2$ scene will be used as the 15 ember dataset in Fig.~\ref{fig:modelresults} and Fig.~\ref{fig:sim+real} for experimentation. The Sunny$^3$ scene dataset will be used as the 50 embers dataset (referred to as the 50embers dataset in the following) in Figs.~\ref{fig:modelresults} and \ref{fig:sim+real}. Finally, the Mix Skylight/Temperature$^4$ scene dataset will participate in the evaluation as the 200 ember dataset (referred to as the 200embers dataset) in Figs.~\ref{fig:modelresults} and \ref{fig:sim+real}.

%% file: experimentalEva.tex
\section{Experimental Evaluation} 
\subsection{Experimental Setup}
We performed bounding box detection experiments on four representative subsets of the FireFly dataset. 
Four models were evaluated in this process: Sparse R-CNN, RetinaNet, DETR, and YOLOv7, using default hyperparameters from MMDetection~\cite{chen2019mmdetection} and the YOLOv7 official repository~\cite{wang2022yolov7}. 





Considering that the embers of the 200ember dataset are extremely small and dense, we adjusted the pre-processing of the image to have size $1,280 \times 1,280$ to obtain higher resolution features. 

When testing the performance of different models on the same dataset and evaluating the performance of detection models trained on various datasets, we employ a metric known as mAP, or 'mean Average Precision.' As depicted in Figs.~\ref{fig:modelresults} and ~\ref{fig:sim+real}, we calculate the model's average detection accuracy at an IOU threshold of 0.5 and from 0.5 to 0.95. These metrics allow us to effectively assess the model's capacity to detect objects accurately within the given datasets and comprehend its robustness across different levels of spatial precision.

\subsection{Comparative Analysis of Detection Models}
\begin{figure}[!t]
	\centering
	\begin{minipage}{\linewidth}
		\centering
		\includegraphics[width=\textwidth]{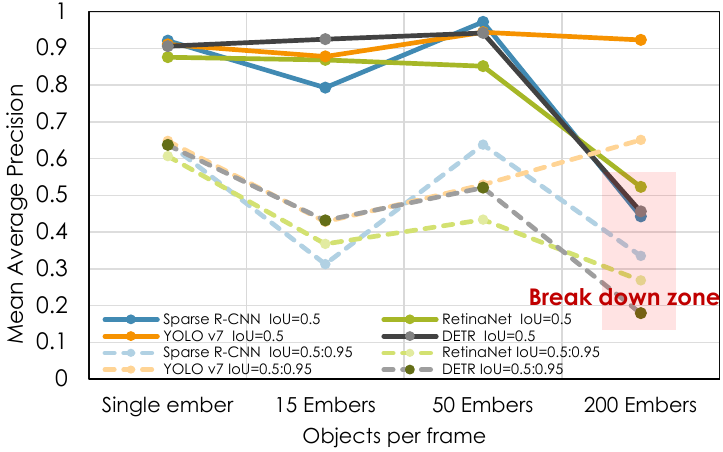}
		   \caption{Detection comparison among models}
		\label{fig:modelresults}
	\end{minipage}
\end{figure}

From the results shown in Fig.~\ref{fig:modelresults}, we can see that all 4 models perform well on the identified synthetic subsets of 1 ember, 15 embers and 50 embers per frame. However, with the increase in the number of embers and the reduction in size, the metrics of RetinaNet, Sparse R-CNN and DETR all dropped sharply. The mAP for DETR plummeted from 0.94 to 0.45, RetinaNet descended from 0.85 to 0.52, and Sparse R-CNN dropped from 0.97 to 0.44. This is mainly due to the backbone resolution they use, which makes it difficult to learn the features of extremely small targets in high-resolution images. Whether it is downsampling achieved by convolution or pooling, a large amount of target information is lost. At the same time, it can be observed that the attention mechanism in DETR may tend to focus more on global features, which could potentially create certain challenges for the detection of small targets. However, the network of YOLOv7 is designed based on high-resolution input and has the best adaptability to our task. We thus choose YOLOv7 for subsequent experiments in this paper.

\subsection{Evaluations in Real Ember Applications}
\begin{figure}
    \centering
    \includegraphics[width=0.45\textwidth]{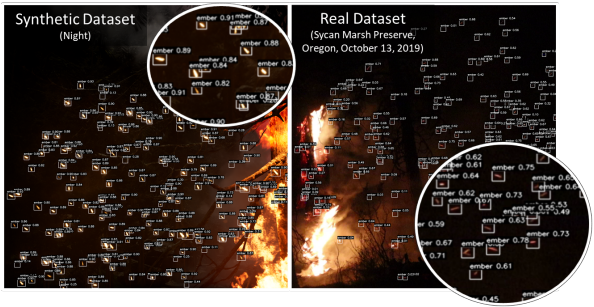}
    \caption{Real wildfire sparks frame evaluation}
    \label{fig:real}
\end{figure}
To test our framework on real data, we used unlabeled frames from forest fire videos acquired from U.S. Department of Agriculture that contain embers of a wildfire at Sycan Marsh Preserve, Oregon~\cite{Jim}. Since the amount of embers produced by fires in real scenes is often very large and densely distributed, as illustrated on the right-side of Fig.~\ref{fig:real}, the cost of manually labeling these frames is very high. In order to mitigate this problem, we developed an 
automatic labeling program that leverages the YOLOv7 model~\cite{wang2022yolov7} in the AutoLabelImg framework~\cite{AutoLabelImg}. We first used our YOLOv7 model trained on our synthetic dataset to perform automatic labeling and then applied manual adjustments to correct for bad/missing bounding boxes. The result is a small semi-automatically-labelled real-world dataset containing a total of 240 images, of which 210 are used as training set and 30 are used for evaluation.

\begin{figure}[!t]
	\centering
	\begin{minipage}{\linewidth}
		\centering
		\includegraphics[width=\textwidth]{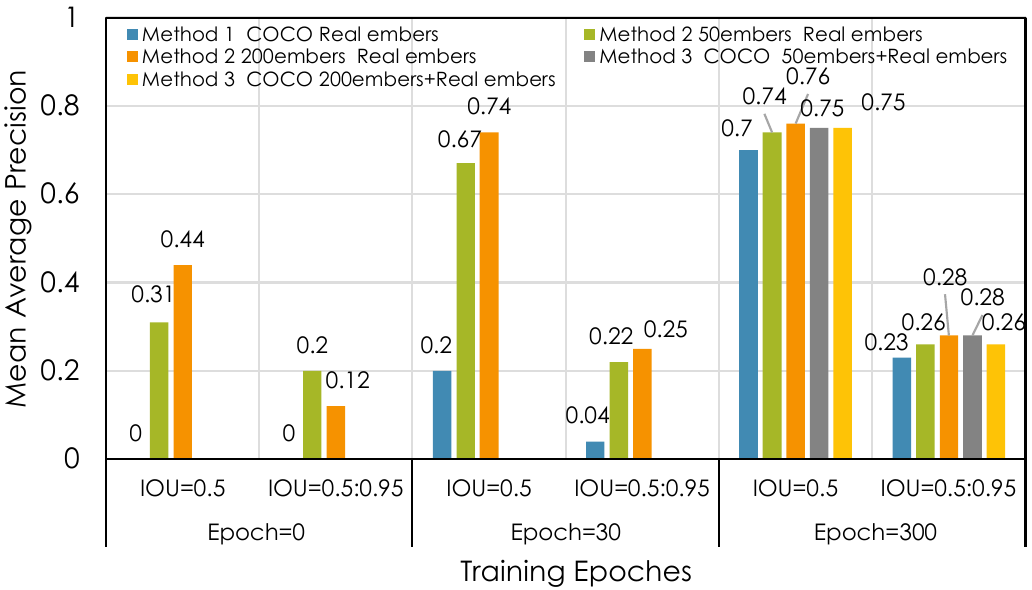}
		   \caption{Comparison of test results on real dataset using different training strategies}
		\label{fig:sim+real}
	\end{minipage}
\end{figure}

Referring to Fig.~\ref{fig:sim+real}, we initially evaluated the performance of the YOLOv7 model trained exclusively on the synthetic dataset by testing it on the real dataset. The results indicated that, despite the absence of any training on the real dataset, the model's performance reached a functional level. Specifically, the model trained on the 50embers dataset achieved a mAP of 0.31 at an $IoU=0.5$ on the real dataset. Similarly, the model trained on the 200embers dataset, as shown in Fig.~\ref{fig:real} left, achieved an mAP of 0.44 at the same IoU on the real dataset. It is noteworthy that the YOLOv7 model trained on the COCO dataset yielded a test metric of zero on the real dataset, underscoring the necessity and utility of our dataset for this task.


Our investigation encompassed three distinct training methodologies. Method 1, denoted by the blue label in Fig.~\ref{fig:sim+real}, involved employing the actual dataset for comprehensive supervised learning, leveraging the COCO dataset~\cite{lin2014microsoft} for pretraining, and then fine-tuning with real ember datasets.

Method 2, depicted by the green label (for 50 embers) and orange label (for 200 embers) in Fig.~\ref{fig:sim+real}, pretrains the model on synthetic datasets and fine-tunes it using real ember datasets, serving as a foundation for implementing transfer learning on real-world datasets.

Lastly, Method 3, a hybrid strategy, utilizes a combination of synthetic and real-world datasets for comprehensive supervised learning. Here, we employed the COCO dataset for pretraining and a hybrid dataset comprised of the 50 embers (gray label in Fig.~\ref{fig:sim+real}) and 200 embers (yellow label in Fig.~\ref{fig:sim+real}) datasets, supplemented with real ember datasets, for fine-tuning.

The mAP at $IoU=0.5$ for the model trained using Method 1 converges to 0.70. This value is readily surpassed by Method 2 after 30 epochs of training. The mAP at $IoU=0.5$ for the model, which leverages pre-trained weights on the 200embers dataset, reaches 0.74. The final optimal mAP for Method 2 is 0.76, representing a 8.57\% improvement over Method 1. Among all experiments, Method 3 achieves the highest score of 0.28 on the more stringent mAP metric with an IoU range of 0.5 to 0.95. Method 2 is our recommended approach due to its efficient training, rapid convergence, superior performance, and all without the need for additional manual dataset mixing.

%% file: conclusion.tex
\section{Conclusions} 
We designed a tool to automatically generate synthetic labeled dataset with dense small targets, and made the parameters controlling the diversity of the dataset adjustable based on UE. Using this approach, combined with reconstructions of forest scenes and simulations of ember trajectories, we released the large-scale synthetic dataset FireFly for ember detection. Our dataset consists of 19,273 images with high diversity. We tested the dataset on popular object detection models and demonstrates the relative effectiveness of YOLOv7. Using our training strategies for the real wildfire scene, the optimal mAP reaches 0.76 at $IoU=0.5$, and 0.28 at $IoU=0.5:0.95$, which is 8.57\% higher than training solely on a small semi-automatically-labelled real dataset. In addition, our training strategy achieves higher accuracy with fewer epochs than training only with the real dataset. These results suggest our tool and dataset provide an effective framework for ML-based detection of real-life wildfire embers.